%% file: main.tex
\documentclass[10pt,twocolumn,letterpaper]{article}

\usepackage[review]{iccv}      

\input{preamble}

\definecolor{iccvblue}{rgb}{0.21,0.49,0.74}
\usepackage[pagebackref,breaklinks,colorlinks,allcolors=iccvblue]{hyperref}

\def\paperID{*****} 
\def\confName{ICCV}
\def\confYear{2025}

\title{Adversarial Data Augmentation for Single Domain Generalization via Lyapunov Exponent-Guided Optimization}

\author{
Zuyu Zhang, Ning Chen, Yongshan Liu, Qinghua Zhang, Xu Zhang\thanks{Corresponding author.}\\
Chongqing University of Posts and Telecommunications\\
{\tt\small \{zhangzuyu, s230233004, liuyongshan,zhangqh, zhangx\}@cqupt.edu.cn}
}

\begin{document}
\maketitle
\input{sec/0_abstract}    
\input{sec/1_intro}
\input{sec/2_relatedwork}
\input{sec/3_methodology}
\input{sec/4_experiments}

\input{sec/5_conclusion}
{
    \small
    \bibliographystyle{ieeenat_fullname}
    \bibliography{main}
}


\end{document}

%% file: preamble.tex
\usepackage{amsmath}    
\usepackage{amssymb}    
\usepackage{algorithm}  
\usepackage{algpseudocode} 
\usepackage{graphicx}   
\usepackage{subcaption}
\usepackage{threeparttable}
\usepackage{makecell}

%% file: sec/0_abstract.tex
\begin{abstract}


Single Domain Generalization (SDG) aims to develop models capable of generalizing to unseen target domains using only one source domain, a task complicated by substantial domain shifts and limited data diversity. Existing SDG approaches primarily rely on data augmentation techniques, which struggle to effectively adapt training dynamics to accommodate large domain shifts. To address this, we propose LEAwareSGD, a novel Lyapunov Exponent (LE)-guided optimization approach inspired by dynamical systems theory. By leveraging LE measurements to modulate the learning rate, LEAwareSGD encourages model training near the edge of chaos, a critical state that optimally balances stability and adaptability. This dynamic adjustment allows the model to explore a wider parameter space and capture more generalizable features, ultimately enhancing the model's generalization capability. Extensive experiments on PACS, OfficeHome, and DomainNet demonstrate that LEAwareSGD yields substantial generalization gains, achieving up to 9.47\% improvement on PACS in low-data regimes. These results underscore the effectiveness of training near the edge of chaos for enhancing model generalization capability in SDG tasks.
\end{abstract}

%% file: sec/1_intro.tex
\section{Introduction}
\label{sec:intro}

Deep neural networks have shown remarkable progress across various tasks in standard supervised learning settings where the training and test data come from the same distribution \cite{he2016deep,li2019deep}. However, their performance often suffers significantly when encountering domain shift, a prevalent issue in real-world applications \cite{pan2009survey,shao2014transfer,zhuang2020comprehensive}. For instance, a model trained on DSLR images may struggle when tested on smartphone images due to differences in resolution, noise, and other factors. These domain shifts are pervasive in real-world scenarios, emphasizing the need for models that can generalize effectively to unseen distributions.

\begin{figure}[htbp]
    \centering
    \begin{subfigure}{0.23\textwidth}
        \centering
        \includegraphics[width=\linewidth]{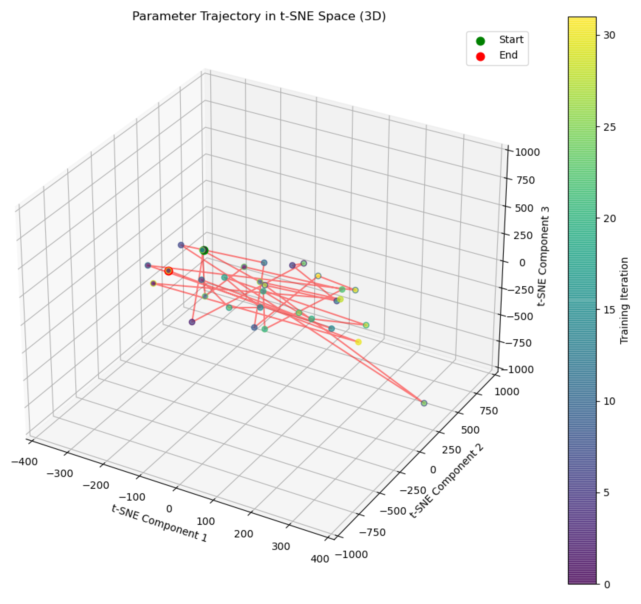} 
        \caption{ADA}
        \label{fig:sub1}
    \end{subfigure}
    \hfill
    \begin{subfigure}{0.23\textwidth}
        \centering
        \includegraphics[width=\linewidth]{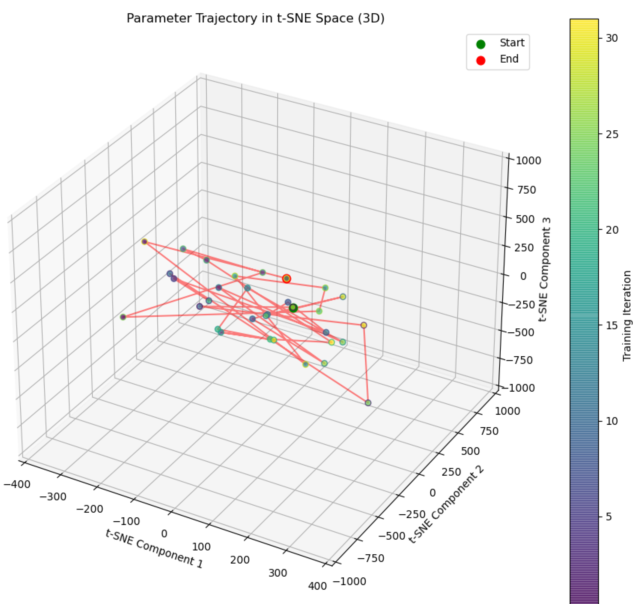} 
        \caption{ME-ADA}
        \label{fig:sub2}
    \end{subfigure}
    
    \vspace{0.5cm} 

    \begin{subfigure}{0.23\textwidth}
        \centering
        \includegraphics[width=\linewidth]{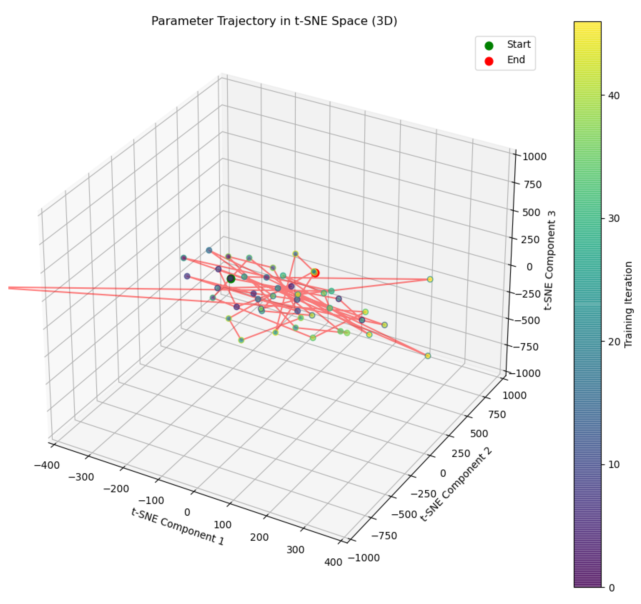} 
        \caption{AdvST}
        \label{fig:sub3}
    \end{subfigure}
    \hfill
    \begin{subfigure}{0.23\textwidth}
        \centering
        \includegraphics[width=\linewidth]{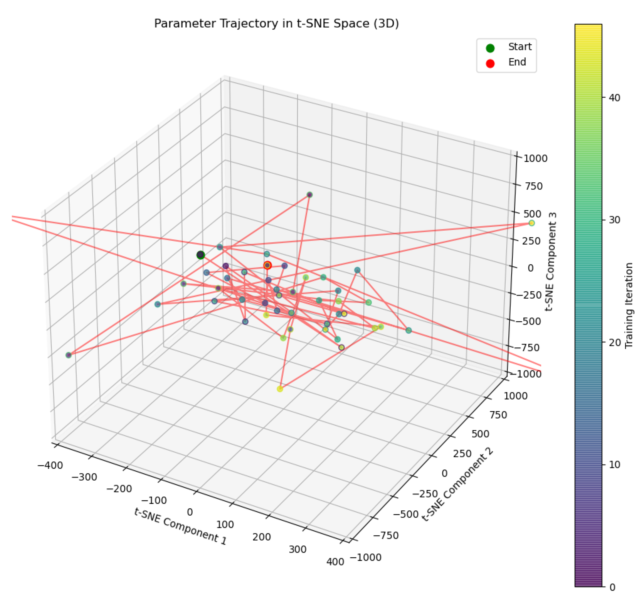} 
        \caption{Ours}
        \label{fig:sub4}
    \end{subfigure}
    
    \caption{t-SNE visualization of parameter trajectories in initial epochs for a model trained on the PACS dataset \cite{li2017deeper}. Each point represents the model’s parameters after an iteration. Comparison between (a) ADA \cite{volpi2018generalizing}, (b) ME-ADA \cite{zhao2020maximum}, (c) AdvST \cite{zheng2024advst}, and (d) our method demonstrates that our approach can explore a broader parameter space, facilitating the learning of more generalizable features across unseen domains.}

    \label{fig:intro}
\end{figure}

Single Domain Generalization (SDG) methods aim to address these domain shifts when only a single source domain is available for training. Current SDG approaches primarily employ data augmentation techniques to generate diverse training samples \cite{volpi2018generalizing,zhao2020maximum,qiao2020learning,li2021progressive,xu2023simde,yang2024causality}. Methods such as adversarial data augmentation \cite{volpi2018generalizing,zhao2020maximum,qiao2020learning} generate perturbed samples that simulate potential domain shifts. 
Despite their contributions, as shown in Figure \ref{fig:intro}, these adversarial data augmentation approaches often fail to adequately capture the global structure of domain shifts in the parameter space due to their emphasis on localized perturbations. As a result, they may not fully explore the parameter space, limiting the model's capability to generalize across diverse distributions. 

In this work, we draw inspiration from dynamical systems theory \cite{alligood1998chaos,feng2019optimal} and propose a novel perspective of model perturbations based on the edge of chaos. This critical transition point maintains a balance between order and chaos in dynamical systems. In SDG, training models near the edge of chaos can help capture features that better generalize across domains by maintaining both stability and adaptability \cite{zhang2021edge}. 
To quantitatively measure a model’s degree of chaos, we employ the Lyapunov Exponent (LE) \cite{hayashi2024chaotic} as a metric to estimate the model's proximity to this edge of chaos. A high LE indicates chaotic behavior and sensitivity to perturbations, whereas a low LE suggests stable dynamics that may restrict the model's generalization capability. 
We frame the training of neural networks as a discrete-time dynamical system in the parameter space, where each parameter update represents a state transition in this system. By dynamically adjusting the LE, we aim to guide adversarial data augmentation in a way that encourages broader exploration of the parameter space, achieving a balance between stability and adaptability for improved domain generalization.

Building on this insight, we introduce LE-Aware Stochastic Gradient Descent (LEAwareSGD), a novel optimizer that dynamically adjusts the learning rate based on LE computations. 
By integrating adversarial data augmentation with LE-aware optimization, LEAwareSGD generates diverse perturbations that enhance the model’s ability to generalize to unseen domains while guiding it to operate near the edge of chaos. 
Unlike conventional optimizers \cite{robbins1951stochastic,kingma2014adam,loshchilov2017decoupled,tieleman2012rmsprop} without considering the system's state, LEAwareSGD leverages LE as a feedback mechanism, adjusting the learning rate in response to the system's state and enabling the model to explore the parameter space effectively for capturing more generalizable features. 

Our contributions are summarized as follows:
\begin{itemize} [noitemsep, topsep=0pt, partopsep=0pt, parsep=0pt]
    \item We conceptualize the training of neural networks as a discrete-time dynamical system and leverage the LE to assess the system's proximity to the edge of chaos.
    \item We propose LEAwareSGD, an optimizer for adversarial data augmentation that dynamically adjusts the learning rate based on LE, guiding the training process near the edge of chaos to enhance generalization.
    \item Extensive experiments on three standard SDG benchmarks, including PACS \cite{li2017deeper}, OfficeHome \cite{venkateswara2017deep}, and DomainNet \cite{peng2019moment}, validate the effectiveness of the proposed LEAwareSGD to improve cross-domain generalization. 
\end{itemize}

%% file: sec/2_relatedwork.tex
\section{Related Work}

\subsection{Single Domain Generalization}

Single domain generalization seeks to improve model robustness by enabling generalization from a single source domain to unseen target domains, eliminating the need for target data during training. Most SDG methods rely on data augmentation \cite{zhang2020generalizing, qiao2020learning} to increase diversity within the source domain, while generative model-based approaches \cite{li2021progressive, xu2023simde, yang2024causality} synthesize samples to extend the source distribution. Although these generative techniques introduce sample variety, they are often computationally demanding and complex to train, frequently producing samples that lack sufficient quality and diversity. This limitation can cause models to overfit to synthetic data rather than enhancing generalization to real-world domains.
Adversarial data augmentation \cite{zhao2020maximum, cheng2023adversarial, zheng2024advst} offers an alternative by generating adversarial examples that mimic domain shifts to improve model robustness. However, these examples often introduce only localized perturbations, failing to explore the parameter space extensively. 
In contrast, our proposed LE-guided optimization introduces a dynamic SDG framework that modulates model sensitivity to perturbations. This LE-aware optimization promotes broader exploration of the parameter space, effectively capturing complex domain shifts and improving cross-domain generalization capability.

\subsection{Optimizers in Domain Generalization} 
Optimizers play a critical role in model generalization, with different strategies yielding varied outcomes depending on the loss landscape's properties. Widely-used optimizers like Stochastic Gradient Descent (SGD) \cite{nesterov1983method} and Adam \cite{kingma2014adam} are commonly used in deep learning tasks, yet they exhibit different behaviors in terms of generalization. Some prior works \cite{hardt2016train,zhang2023flatness} show that Adam often results in worse generalization than SGD due to its susceptibility to sharp minima. In contrast, SGD tends to find flatter minima more conducive to robust generalization. This distinction highlights the importance of the loss surface's geometry, as flatter minima are generally associated with improved generalization performance, especially in out-of-distribution settings.
In light of this, recent optimization strategies have been explicitly designed to address the sharpness of minima. Sharpness-Aware Minimization (SAM) \cite{foret2020sharpness} represents a key development in this direction, where the objective is to minimize sharpness and seek flatter regions of the loss landscape. SAM and its variants \cite{zhuang2022surrogate, zhang2023flatness} have demonstrated notable improvements in generalization, particularly in distribution. Recent improvements such as GSAM \cite{zhuang2022surrogate}, which introduces a surrogate gap to better distinguish between sharp and flat minima, further enhance generalization by guiding the model toward regions with both low sharpness and low loss. 
Our proposed LEAwareSGD differs from these optimizers by dynamically adjusting the learning rate based on the LE, allowing the model toward the edge of chaos and exploring a broader parameter space for improving generalization capability.

\subsection{Dynamical Systems in Deep Learning} 
The application of dynamical systems theory to deep learning has introduced new perspectives on neural network training, particularly in understanding stability, and generalization \cite{liu2019deep,tang2020introduction}. 
By modeling neural networks as discrete-time systems, researchers have used concepts such as chaos theory to enhance generalization, especially within residual networks via flow maps \cite{li2022deep}. This perspective underscores the significance of training near the edge of chaos, a state that balances stability and adaptability \cite{zhang2021edge, zhang2024asymptotic}.
Lyapunov exponent (LE) is a key tool to quantitatively measure the model's training stability, particularly in recurrent architectures \cite{vogt2022lyapunov, guarneros2022lyapunov}. However, existing methods have not fully leveraged LE for direct control of training under domain shifts, particularly in the context of adversarial data augmentation for domain generalization tasks. Our proposed LEAwareSGD  addresses this limitation by using LE to dynamically guide the optimization process, effectively enhancing model generalization capability.

%% file: sec/3_methodology.tex
\section{Methodology}

\subsection{Overview of LEAwareSGD}
We propose LEAwareSGD, a novel optimization method for adversarial data augmentation to enhance generalization by exploiting the edge of chaos. In this approach, the LE serves as a metric to identify and explore regions in the parameter space where generalizable features are more likely to emerge.



\subsection{LE-Based Model Perturbation}
The LE quantifies the rate at which small perturbations in model parameters evolve during training, offering insights into the stability of the optimization process. Here, we denote the model parameters $\theta_t$ at iteration $t$ and the standard gradient descent update can be formulated as follows:

\vspace{-5pt}
\begin{equation}
\theta_{t+1} = \theta_t - \eta_t \nabla L(\theta_t),
\label{eq.1}
\end{equation}
where $\eta_t$ denotes the learning rate and $L(\theta_t)$ represents the loss function.

To quantify how perturbations evolve during training, we introduce an initial perturbation $\delta \theta_0$ to the model parameters.Parameters after perturbation is $\widetilde{\theta_t}:=\theta_t+\delta\theta_t$,where $\delta $ is a tiny real-valued scalar hyperparameter.The perturbation at step $t+1$ propagates according to the following formula:

\vspace{-5pt}
\begin{equation}
\widetilde{\theta}_{t+1}=\widetilde{\theta_t}-\eta_t\nabla L(\widetilde{\theta_t}).
\label{eq.2}
\end{equation}

Combining Equation \ref{eq.1} and Equation \ref{eq.2}, we get the following update for the perturbation:

\vspace{-5pt}
\begin{equation}
\delta\theta_{t+1}=\delta\theta_t-\eta_t\left(\nabla L(\widetilde{\theta_t})-\nabla L(\theta_t)\right).
\label{eq.3}
\end{equation}

To further analyze the propagation of this perturbation, we perform a first-order Taylor expansion of the loss function $L(\widetilde{\theta_t})$ at $\theta_t$ and calculate the gradient:

\vspace{-5pt}
\begin{equation}
\nabla L(\widetilde{\theta_t})=\nabla L(\theta_t)+H[L(\theta_t)]\delta\theta_t+o({\|\delta\theta_t\|}^2),
\label{eq.4}
\end{equation}
where $H[L(\theta_t)]$ is the Hessian matrix, and $o(\|\delta\theta_t\|^2)$ represents higher-order terms. Ignoring these higher-order terms, and substituting Equation \ref{eq.4} into Equation \ref{eq.3}, we approximate the evolution of the perturbation as:

\vspace{-5pt}
\begin{equation}
\delta \theta_{t+1} = \left( I - \eta_t H[L(\theta_t)] \right) \delta \theta_t.
\end{equation}

By recursively applying this relationship, we obtain the propagation of the perturbation over time:

\vspace{-5pt}
\begin{equation}
\begin{aligned}
\delta\theta_t&=\left(I-\eta_{t-1} H[L(\theta_{t-1})]\right)\left(I-\eta_{t-2} H[L(\theta_{t-2})]\right)\\&\cdots\left(I-\eta_0 H[L(\theta_0)]\right)\delta\theta_0.
\label{eq.6}
\end{aligned}
\end{equation}

The definition of the Lyapunov exponent is:
\begin{equation}
\text{LE} = \lim_{t \to \infty} \frac{1}{t} \ln \left( \frac{ \| \delta \theta_t \| }{ \| \delta \theta_0 \| } \right) 
\label{eq.7}
\end{equation}

 By substituting Equation \ref{eq.6} into the definition of LE, we can derive the connection between the model's learning rate $\eta$ and LE:


\vspace{-5pt}
\begin{equation}
\begin{aligned}
\text{LE}&\geq\lim_{t\to\infty}\frac{1}{t}\sum_{i=0}^{t-1}\ln(1-\eta_i\| H[L(\theta_i)]\|)\\\text{LE}&\leq\lim_{t \to \infty} \frac{1}{t} \sum_{i=0}^{t-1} \ln \left( \left\| I - \eta_i H[L(\theta_i)] \right\| \right),
\end{aligned}
\end{equation}
where the LE is determined by the learning rate $\eta_i$ and the Hessian matrix $H[L(\theta_i)]$. 


\subsection{LE-Guided Optimization}
A positive LE indicates the exponential growth of perturbations, signifying instability, while a negative LE suggests convergence and stability. However, extreme stability or instability can lead to either overfitting or convergence difficulties. 
Our approach aims to guide model parameters toward an LE value close to zero but slightly negative, positioning the system near the edge of chaos. This critical state balances stability with adaptability, enhancing the model’s generalization capacity. To achieve this, we dynamically adjust the learning rate to amplify perturbations, thereby steering the LE toward closer to the edge of chaos.
Specifically, the innovation of the proposed LEAwareSGD lies in its learning rate adjustment based on LE dynamics, allowing the model to adapt effectively across training phases. We modulate the learning rate according to changes in the LE, $\Delta \text{LE}_t = \text{LE}_t - \text{LE}_{t-1}$. 

\vspace{-10pt} 
\begin{equation}
\eta_{t+1} = \eta_t \cdot \exp\left( -\beta \cdot \Delta \text{LE}_t \right) \quad \text{if} \quad \Delta \text{LE}_t > 0
\label{eq.9}
\end{equation}

\noindent where $\beta$ is a weighting parameter that controls the sensitivity of the learning rate adjustment. When $\Delta \text{LE}_t > 0$, indicating that the model is approaching regions near the edge of chaos where generalizable features are more likely to emerge, we reduce the learning rate to allow for deeper exploration of these regions. This approach ensures that the model explores parameter regions with strong generalization potential while avoiding over-stabilization that could hinder the learning process. When $\Delta \text{LE}_t \leq 0$, we leave the learning rate unchanged, maintaining the current exploration rate.

\subsection{LE-Based Adversarial Data Augmentation}
To further enhance the model's robustness to domain shifts, we integrate LEAwareSGD with adversarial data augmentation. Adversarial data augmentation generates samples that simulate domain shifts, exposing the model to a broader range of data variations.
Given a model $f_\theta$ with parameters $\theta$, we define adversarial samples as transformations of the original data. Let $\tau(x; \omega)$ denote a set of semantic transformations parameterized by $\omega$. The joint optimization objective is: 


\vspace{-10pt}
\begin{equation}
\begin{aligned}
\min_{\theta} \max_{\omega} \mathbb{E}_{(x, y) \sim \mathcal{D}_S} & \left[ \ell\left( \theta; \tau(x; \omega), y \right) - \lambda d_\theta\left( \tau(x; \omega), x \right) \right] \\
&+ \frac{\gamma}{2} \cdot \|\theta\|_2^2,
\end{aligned}
\label{eq.ada}
\end{equation}

\noindent where $\ell(\theta; \cdot)$  is the prediction loss, $\lambda$ balances the adversarial loss and feature consistency, and  $d_\theta(\cdot)$ represents the feature distance between the original and transformed samples. Here, the regularization term $\frac{\gamma}{2} \cdot \|\theta\|_2^2$ denotes a weight decay, ensuring that the Hessian matrix remains approximately positive definite. This regularization encourages LE to be negative, facilitating the stability of model training.
Note that this adversarial training procedure is guided by LE-based optimization, which encourages the model to generalize to unseen domains by learning robust features invariant to domain-specific perturbations.


\subsection{Training Procedure}
The training procedure involves iterative steps alternating between adversarial example generation and parameter updates using LEAwareSGD. In each epoch, adversarial examples are first generated to simulate unseen domain shifts, fostering the model to learn invariant features. Subsequently, LEAwareSGD adjusts the model's parameters with a LE-guided learning rate to encourage the parameters' updating towards the edge of chaos. This LE-guided adaptation encourages the exploration of parameter regions conducive to generalization while controlling sensitivity to perturbations. 
The complete algorithm is shown in Algorithm \ref{alg:leaware_sgd}. We also provide a detailed theoretical analysis of the proposed LEAwareSGD in Suppl. Sec-\ref{supp1.1}.

\begin{algorithm}[H]
\caption{LEAwareSGD}
\label{alg:leaware_sgd}
\begin{algorithmic}[1]
\State \textbf{Input:} Initial parameters $\theta_0$, learning rate $\eta_0$, weighting parameter $\beta$, number of iterations $N$
\State \textbf{Output:} Optimized parameters $\theta^*$
\For{$t = 0, 1, 2, \dots, N-1$}
    \State Compute gradient $\nabla_\theta L(\theta_t)$
    \State Update parameters: $\theta_{t+1} = \theta_t - \eta_t \nabla_\theta L(\theta_t)$
    \State Compute updated model perturbation using Eq. \ref{eq.6}
    \State Calculate LE using Eq. \ref{eq.7}
    \State Update $\Delta \text{LE}_t = \text{LE}_t - \text{LE}_{t-1}$
    \If{$\Delta \text{LE}_t > 0$} 
        \State $\eta_{t+1} = \eta_t \cdot \exp\left( -\beta \cdot \Delta \text{LE}_t \right)$ 
    \Else
        \State $\eta_{t+1} = \eta_t$
    \EndIf
\EndFor
\State \textbf{Return:} $\theta^*$
\end{algorithmic}
\end{algorithm}



%% file: sec/4_experiments.tex
\section{Experiment}
\subsection{Datasets and Settings}

\textbf{Datasets.} 
We evaluate our method on three widely-used domain generalization benchmarks: PACS \cite{li2017deeper}, OfficeHome \cite{venkateswara2017deep}, and DomainNet \cite{peng2019moment}, arranging them in increasing levels of difficulty to comprehensively test the model’s robustness across varying domain shifts. \textbf{PACS} consists of four domains, namely, Photo (P), Art Painting (A), Cartoon (C), and Sketch (S), comprising 9,991 images from 7 categories. \textbf{OfficeHome} includes four domain, namely, Art (A), Clipart (C), Product (P), and Real-World (R), with 15,500 images across 65 classes. Compared to PACS, OfficeHome introduces greater diversity in both object categories and domain styles. \textbf{DomainNet} is the most challenging, containing six domains, namely, Real (R), Infograph (I), Clipart (C), Painting (P), Quickdraw (Q), and Sketch (S), with 596,010 images from 345 classes. The substantial domain shifts and the high number of classes in this dataset present considerable challenges for domain generalization.

\textbf{Implementation details.} 
 In all experiments, we employed ResNet-18 \cite{he2016deep} as the backbone model and resized all input images to 224 × 224. The network was trained with LEAwareSGD, setting the momentum to 0.9 and $\beta$ to 1e-1. To ensure a fair comparison, we excluded the test-time learning stage of PSDG\cite{yang2024practical} in all of our experiments. For PACS and OfficeHome datasets, we follow the standard protocol of using one domain as the source domain and the remaining three domains as the target domains \cite{zhao2020maximum,zheng2024advst}. For the PACS dataset, we adopted a batch size of 16, $\gamma$ of 5e-4, and an initial learning rate of 5e-4, training for 50 epochs. For the OfficeHome dataset, we used a batch size of 32, set $\gamma$ to 1e-4, and initialized the learning rate at 1e-4, training for 50 epochs. For experiments conducted on the DomainNet dataset, we follow \cite{xu2023simde, zheng2024advst} and use Real as the source domain and the remaining five domains as the target domains. The batch size was increased to 128, $\gamma$ was set to 1e-5, and the initial learning rate was set to 1e-3, with a training duration of 200 epochs. For hyperparameter tuning, we fix all training settings (e.g., random seed, learning rate) and tune only $\beta$ and $\gamma$, with the best setting shown in Fig.\ref{fig_parameters}. All experiments were performed on an NVIDIA A100 GPU, and each experiment was repeated 10 times with different random seeds. The results are reported as the average accuracy alongside the standard deviation.

\subsection{Comparison with SDG Approaches}

\textbf{Results on PACS.} 
This adjustment aligns the evaluation setup with our method, which operates solely during the training phase. The evaluation results on the PACS dataset, presented in Table \ref{tab_pacs}, indicate that our method achieves the highest average accuracy of 69.46\%, outperforming all compared methods. Specifically, our model achieves the highest performance gains in both the A and P domains, with classification accuracies of 79.17\% and 65.05\%, respectively. 
Our approach outperforms the second-best method, PSDG (67.14\%), with an improvement of 2.32\% in average accuracy. This result underscores the effectiveness of our LE-guided adversarial data augmentation strategy, which enhances model generalization by facilitating the exploration of a broader parameter space and reducing overfitting to domain-specific features.


\begin{table}[htbp]
    \centering
    \setlength{\tabcolsep}{5pt}
     \small
    \begin{tabular}{lccccc}
    \toprule
        Methods & A & C & P & S & Avg. \\ \midrule
        ERM & 70.49 & 73.56 & 45.92 & 41.21 & 57.80  \\
        MixUp  \cite{yan2020improve} & 67.60  & 67.13  & 51.59  & 47.04  & 58.34  \\
        RSC \cite{huang2020self} & 73.40  & 75.90  & 56.20  & 41.60  & 61.78  \\
        ADA \cite{volpi2018generalizing} & 71.56  & 76.84  & 52.36  & 43.66  & 61.11  \\
        ME-ADA \cite{zhao2020maximum} & 71.52  & 76.83  & 46.22  & 46.32  & 60.22  \\
        SAGM \cite{wang2023sharpness} & 65.94  & 70.10  & 42.14  & 53.65  & 57.96  \\
        RandConv \cite{xu2020robust} & 73.68  & 74.88  & 55.42  & 46.77  & 62.69  \\
        MetaCNN \cite{wan2022meta} & 76.50  & 74.50  & 56.70  & 55.70  & 65.80  \\
        L2D \cite{wang2021learning} & 77.00  & 76.55  & 53.22  & 54.44  & 65.30  \\
        PDEN \cite{li2021progressive} & 76.43  & 73.87  & 58.52  & 53.92  & 65.69  \\
        PSDG  \cite{yang2024practical} &  77.88  & 75.24  & 55.49  & \textbf{59.96}  & 67.14  \\
        AdvST \cite{zheng2024advst}  & 76.65  & 74.92  & 62.47  & 54.18  & 67.06  \\
        Ours & \textbf{79.17} & \textbf{77.16} & \textbf{65.05} & 57.78 & \textbf{69.46}\scriptsize{$\pm$0.33} \\ \bottomrule
    \end{tabular}
    \caption{Classification accuracy (\%) comparison on the PACS dataset. The best results are in bold.}
    \label{tab_pacs}
\end{table}

\textbf{Results on OfficeHome.} 
Table \ref{tab_office} shows the classification accuracy comparison of our method with existing baselines on the OfficeHome dataset. Our proposed method consistently outperforms other approaches, achieving the highest accuracy across all domains (A, C, P, and R) with an average accuracy of 54.38\%. Notably, our method surpasses the previous best method, AdvST, which obtained an average accuracy of 52.60\%, with an improvement of 1.78\%. This gain highlights the effectiveness of our LE-guided optimization, which facilitates the exploration of a broader parameter space and enhances the model's ability to generalize to unseen domains.

\begin{table}[htbp]
    \centering
    \setlength{\tabcolsep}{5pt}
    
    \small
    \begin{tabular}{lccccc}
    \toprule
        Methods & A & C & P & R & Avg. \\ \midrule
        ERM & 42.74  & 39.31  & 39.86  & 52.50  & 43.60  \\
        MixUp  \cite{yan2020improve} & 43.69  & 40.68  & 38.56  & 52.12  & 43.76  \\
        SagNet \cite{nam2021reducing} & 49.84  & 43.01  & 41.42  & 55.61  & 47.47  \\
        RandConv \cite{xu2020robust} & 45.79  & 40.20  & 37.45  & 52.87  & 44.08  \\
        Fish \cite{shi2021gradient}  & 45.10  & 39.74  & 36.67  & 52.09  & 43.40  \\
        ADA \cite{volpi2018generalizing} & 45.07  & 36.07  & 44.43  & 53.44  & 44.75  \\
        ME-ADA \cite{zhao2020maximum} & 45.27  & 45.79  & 38.65  & 51.69  & 45.35  \\
        GSAM \cite{zhuang2022surrogate} & 28.97  & 41.40  & 41.72  & 51.50  & 40.90  \\
        SAGM \cite{wang2023sharpness} & 47.42  & 43.27  & 41.36  & 54.97  & 46.76  \\
        Mixstyle \cite{zhou2024mixstyle} & 51.19  & 48.73  & 46.85  & 55.88  & 50.66  \\
        PCGrad \cite{yu2020gradient} & 49.91  & 46.61  & 46.11  & 56.31  & 49.74  \\
        PSDG \cite{yang2024practical}  & 42.80  & 46.40  & 44.31  & 54.68  & 47.05  \\
        AdvST \cite{zheng2024advst} & 51.32  & 52.08  & 48.89  & 58.11  & 52.60  \\
        Ours & \textbf{52.89} & \textbf{55.18} & \textbf{51.69} & \textbf{58.95} & \textbf{54.38}\scriptsize{$\pm$0.30}  \\ 

    \bottomrule
    \end{tabular}
    \caption{Classification accuracy (\%) comparison on the OfficeHome dataset. The best results are in bold.}
    \vspace{-10pt}
    \label{tab_office}
\end{table}

\textbf{Results on DomainNet.} 
Table \ref{tab_domain} shows the classification accuracy comparison on the DomainNet dataset. Our proposed method achieves the highest average accuracy of 28.15\%, surpassing all other methods, including the second-best method, AdvST, which reaches an average accuracy of 27.22\%. Notably, our method performs best in the C and S domains, with accuracies of 44.67\% and 33.74\%, respectively.
In the Q domain, our method achieves an accuracy of 6.70\%, slightly lower than SimDE's 6.85\%. This may be due to the highly abstract nature of the Quickdraw domain, which can benefit from specialized augmentation techniques to better capture its distinct features. Despite this, our method’s overall performance advantage demonstrates the effectiveness of the LE-guided adversarial data augmentation approach in handling the diversity and complexity of DomainNet. 


\begin{table}[htbp]
    \centering
    \setlength{\tabcolsep}{4pt}
    \small
    \begin{tabular}{lcccccc}
    \toprule
        Methods & P & I & C & S & Q & Avg. \\ \midrule
        ERM & 38.05  & 13.31  & 37.89  & 26.26  & 3.36  & 23.77  \\
        MixUp  \cite{yan2020improve} & 38.60  & 13.92  & 38.01  & 26.01  & 3.71  & 24.05  \\
        CutOut \cite{devries2017improved} & 38.30  & 13.70  & 38.40  & 26.20  & 3.70  & 24.06  \\
        AugMix \cite{hendrycks2019augmix} & 40.80  & 13.90  & 41.70  & 29.80  & 6.30  & 26.50  \\
        RandAug \cite{cubuk2020randaugment} & 41.30  & 13.57  & 41.11  & 30.40  & 5.31  & 26.34  \\
        ADA \cite{volpi2018generalizing} & 38.20  & 13.80  & 40.20  & 24.80  & 4.30  & 24.26  \\
        ME-ADA \cite{zhao2020maximum} & 38.48  & 13.56  & 40.66  & 26.05  & 4.42  & 24.63  \\
        ACVC  \cite{cugu2022attention} & 41.30  & 12.90  & 42.80  & 30.90  & 6.60  & 26.90  \\
        PDEN \cite{li2021progressive} & 38.45  & 11.25  & 38.99  & 31.71  & 5.58  & 25.20  \\
        SimDE \cite{xu2023simde} & 39.96  & 12.91  & 41.73  & 33.46  & \textbf{6.85}  & 26.98  \\
        PSDG \cite{yang2024practical} & 41.10  & 13.00  & 41.29  & 31.10  & 4.89  & 26.28 \\
        AdvST \cite{zheng2024advst} & 42.40  & 14.90  & 41.70  & 31.00  & 6.10  & 27.22  \\
        Ours & \textbf{42.77} & \textbf{15.91} & \textbf{44.67} & \textbf{33.74} & 6.70 & \textbf{28.15}\scriptsize{ $\pm$0.61} \\     \bottomrule
    \end{tabular}
    \caption{Classification accuracy (\%) comparison on the DomainNet dataset. The best results are in bold.}
    \vspace{-15pt}
    \label{tab_domain}
\end{table}

\subsection{Comparison with Optimization Approaches}
To further validate the effectiveness of our proposed LEAwareSGD optimizer, we conducted a comparison with four widely used optimization approaches, including Adam  \cite{kingma2014adam}, AdamW \cite{loshchilov2017decoupled}, RMSprop  \cite{tieleman2012rmsprop}, and SGD  \cite{robbins1951stochastic}, on the PACS and OfficeHome datasets. As shown in Table \ref{tab_opt_pacs}, our proposed optimizer achieves the highest average accuracy of 69.46\%, surpassing all other optimizers on the PACS dataset. In contrast, widely used optimizers such as Adam, AdamW, and RMSprop perform significantly worse, with average accuracies of 66.43\%, 66.83\%, and 62.34\%, respectively. This disparity likely stems from the reliance of these adaptive optimizers on variable learning rates, which may cause overfitting to domain-specific features and reduce their generalization ability in domain generalization tasks.
On the other hand, both our LEAwareSGD and standard SGD achieve substantially higher accuracies. While SGD achieves an average accuracy of 67.06\%, our LEAwareSGD optimizer further improves this to 69.46\%. This enhancement underscores the advantage of integrating LE guidance into the optimization process, which enables the model to better capture generalizable features and avoid overfitting to domain-specific characteristics. We present the experimental results on the OfficeHome dataset in Suppl. Sec-\ref{suppC.2}, while analyses of optimizers' performance under different learning rates are provided in Suppl. Sec-\ref{suppC.5}.

\begin{table}[htbp]
    \centering
    \setlength{\tabcolsep}{6pt}
    
    \small
    \begin{tabular}{lccccc}
    \toprule
        Methods & A & C & P & S & Avg. \\ \midrule
        Adam \cite{kingma2014adam} & 76.52  & 71.15  & 64.06  & 53.98  & 66.43  \\
        AdamW \cite{loshchilov2017decoupled} & 76.68  & 71.93  & 62.03  & 56.68  & 66.83  \\
        RMSprop \cite{tieleman2012rmsprop} & 71.62  & 71.08  & 59.09  & 47.55  & 62.34  \\
        SGD \cite{robbins1951stochastic} & 76.65  & 74.92  & 62.47  & 54.18  & 67.06  \\
        Ours & 79.17 & 77.16 & 65.05 & 57.78 & \textbf{69.46}\scriptsize{$\pm$0.33} \\ 
    \bottomrule
    \end{tabular}
    \caption{Comparison results (\%) using different optimizers on the PACS dataset. The best results are in bold.}
    \vspace{-10pt}
    \label{tab_opt_pacs}
\end{table}

    


\subsection{Extension to Multi-Source DG Approaches}
To further validate the scalability of our solution, we perform experiments under the leave-one-out setting of general domain generalization (DG) with ResNet-18 as the backbone, where three domains are combined for training and the remaining domain is reserved for evaluation.Table \ref{tab_MDGpacs} presents the results on the PACS dataset, showing that our method delivers at least comparable performance, highlighting its versatility. This improvement underscores the effectiveness of our LE-guided optimization, which enables a broader exploration of the parameter space and enhances the model’s generalization to unseen domains.

\begin{table}[htbp]
    \centering
    \setlength{\tabcolsep}{5pt}
     \small
    \begin{tabular}{lccccc}
    \toprule
        Methods & A & C & P & S & Avg. \\ \midrule
        RSC \cite{huang2020self} & 82.81  & \textbf{79.74}  & 93.95  & 80.85  & 84.34  \\
        ADA \cite{volpi2018generalizing} & 82.81  & 78.33  & 95.63  & 75.29  & 83.02  \\
        ME-ADA \cite{zhao2020maximum} & 77.88  & 78.58  & 95.33  & 78.07  & 82.47  \\
        Mixstyle \cite{zhou2024mixstyle} & \textbf{83.11}  & 79.43  & \textbf{96.31}  & 72.95  & 82.95  \\
        L2D \cite{wang2021learning} & 81.44  & 79.56  & 95.51  & 80.58  & 84.27  \\
        AdvST \cite{zheng2024advst}  & 78.69  & 72.97  & 91.62  & 80.91  & 81.05  \\
        Ours & 81.35 & 78.88 & 95.45 & \textbf{82.39} & \textbf{84.52} \\ \bottomrule
    \end{tabular}
    \caption{Classification accuracy (\%) comparison for the leave-one-domain-out using the ResNet-18 backbone on the PACS dataset. The column name denotes the test domain, with the other three used for training. The best results are in bold.}
    \vspace{-10pt}
    \label{tab_MDGpacs}
\end{table}

\subsection{Additional Analysis}
\textbf{Performance comparison using different backbone networks.}
To further validate the universality of our solution, we conduct experiments with different ResNet backbones, including ResNet-34, ResNet-50, ResNet-101, and ResNet-152, on the PACS and OfficeHome datasets. 
As shown in Table \ref{tab_backbone_pacs} on the PACS dataset, our approach reaches a performance of  73.68\% on ResNet-34, 71.37\% on ResNet-50, 74.32\% on ResNet-101, and 75.34\% on ResNet-152. These results demonstrate the consistent effectiveness of our method across various backbone architectures, highlighting its superior accuracy, versatility, and robustness. The experimental results for the OfficeHome dataset are presented in Suppl. Sec-\ref{suppC.6}.


\begin{table}[htbp]
    \centering
    \setlength{\tabcolsep}{4pt}
    \small
    \begin{tabular}{llcccccc} 
    \toprule
        Backbone & Methods & A & C & P & S & Avg. \\ \midrule
        ResNet34 
        & PSDG & 79.46 & 80.79 & 57.43 & 58.22 & 68.98& ~ \\
        & AdvST & 79.16 & 76.67 & 65.28 & 60.55 & 70.42& ~ \\
        & Ours & 81.89 & 81.85 & 65.65 & 65.31 & \textbf{73.68}& ~ \\ 
        ResNet50
        & PSDG & 79.85 & 79.72 & 58.08 & 65.68 & 70.83& ~ \\
        & AdvST & 79.65 & 75.05 & 62.62 & 55.75 & 68.27& ~ \\
        & Ours & 81.32 & 78.70 & 65.08 & 60.39 & \textbf{71.37}& ~ \\ 
        ResNet101
        & PSDG & 79.84 & 79.97 & 59.17 & 70.79 & 72.44& ~ \\
        & AdvST & 80.93 & 75.35 & 66.29 & 58.70 & 70.32& ~ \\
        & Ours & 82.95 & 80.79 & 68.57 & 64.95 & \textbf{74.32} & ~ \\ 
        ResNet152
        & PSDG & 82.92 & 81.54 & 62.38 & 71.25 & 74.52& ~ \\
        & AdvST &80.76 & 79.02 & 65.90 & 61.99 & 71.92 & ~ \\
        & Ours & 83.44 & 82.19 & 67.63 & 68.11 & \textbf{75.34} & ~ \\  
    \bottomrule
    \end{tabular}
    \caption{Comparison results (\%) using different ResNet series backbones on the PACS dataset. The best results are in bold.}
    \label{tab_backbone_pacs}
\end{table}



\textbf{Performance comparison with limited training data.}
We compare the performance of our proposed method with AdvST on the PACS and OfficeHome datasets under varying source domain data ratios (10\%, 20\%, and 50\%). As shown in Tables \ref{tab_ratio_pacs} on the PACS dataset, our method consistently outperforms AdvST across all data ratios, demonstrating robust generalization even with limited source domain data for training.
At the 10\% data ratio, our method achieves an average accuracy of 58.73\%, showing a notable improvement of 9.47\% over AdvST’s 49.26\%. With the 20\% data ratio, our approach reaches an average accuracy of 61.78\%, surpassing AdvST by 8.23\%. At the 50\% data ratio, the performance gap narrows slightly, with our method achieving an average accuracy of 66.44\%, which is 6.26\% higher than the 60.18\% achieved by AdvST. The experimental results on the OfficeHome dataset are provided in Suppl. Sec-2.5.
These findings suggest that our method effectively reduces overfitting to specific domains with limited training data, particularly in low-data settings, by encouraging broader exploration of the parameter space to capture generalizable features. 

\begin{table}[htbp]
    \centering
    \setlength{\tabcolsep}{3pt}
    
    \small
    \begin{tabular}{llcccccc} 
    \toprule
        Ratio & Methods & A & C & P & S & Avg. & $\Delta$Avg. \\ \midrule
        10\% & AdvST & 58.74 & 56.94 & 40.62 & 40.73 & 49.26 & ~ \\
        & Ours & 65.43 & 64.43 & 49.40 & 55.66 & 58.73 & \textbf{+9.47} \\ 
        20\% & AdvST & 61.48 & 64.31 & 43.42 & 44.98 & 53.55 & ~ \\
        & Ours & 68.81 & 68.78 & 53.28 & 56.27 & 61.78 & +8.23 \\ 
        50\% & AdvST & 70.27 & 72.74 & 49.52 & 48.20 & 60.18 & ~ \\
        & Ours & 75.88 & 73.55 & 61.60 & 54.72 & 66.44 & +6.26 \\ 
    \bottomrule
    \end{tabular}
    \caption{Performance comparison (\%) using different data ratios on PACS dataset.}
    \vspace{-10pt} 
    \label{tab_ratio_pacs}
\end{table}

    

\textbf{Computational complexity analysis.} To evaluate the computational efficiency of our method, we compared its training time cost with three adversarial data augmentation methods on the PACS and OfficeHome datasets. As shown in Table \ref{tab_timecost}, our approach requires an average training time of 1.99 hours on PACS, slightly higher than AdvST (1.90 hours) but comparable to ADA (2.13 hours) and ME-ADA (2.10 hours). This marginal increase is due to the Lyapunov-guided optimization, which enhances model robustness and generalization. On OfficeHome, our method completes training in 1.19 hours, surpassing AdvST (0.75 hours) but remaining faster than ADA (1.95 hours) and ME-ADA (1.90 hours). While our method introduces a slight time increase compared to AdvST, it achieves significant performance improvements, demonstrating a favorable balance between training time and model performance.


\begin{table}[htbp]
    \centering
    \setlength{\tabcolsep}{6pt}
    
    \small
    \begin{tabular}{lcccccc}
    \toprule
        Dataset & Methods & A & C & P & S & Avg. \\ \midrule
        PACS & ADA & 1.80  & 2.10  & 1.40  & 3.20  & 2.13  \\
        ~ & ME-ADA & 1.80  & 2.10  & 1.30  & 3.20  & 2.10  \\
        ~ & AdvST & 1.58  & 1.87  & 1.40  & 2.76  & 1.90  \\
        ~ & Ours & 1.62  & 2.02  & 1.35  & 2.98  & 1.99  \\ 
        \cmidrule(lr){2-7}
         & Methods & A & C & P & R & Avg. \\ 
        \cmidrule(lr){2-7}
        OfficeHome & ADA & 1.20  & 2.10  & 2.20  & 2.30  & 1.95  \\
        ~ & ME-ADA & 1.20  & 1.90  & 2.20  & 2.30  & 1.90  \\
        ~ & AdvST & 0.75  & 0.74  & 0.75  & 0.76  & 0.75  \\
        ~ & Ours & 1.10  & 1.17  & 1.23  & 1.27  & 1.19  \\ 
    \bottomrule
    \end{tabular}
    \caption{Training time (hours) on PACS and OfficeHome datasets.}
    \label{tab_timecost}
\end{table}

\begin{figure}[htbp]
\vspace{-10pt}
    \centering
    \begin{subfigure}{0.23\textwidth}
        \centering
        \includegraphics[width=\textwidth]{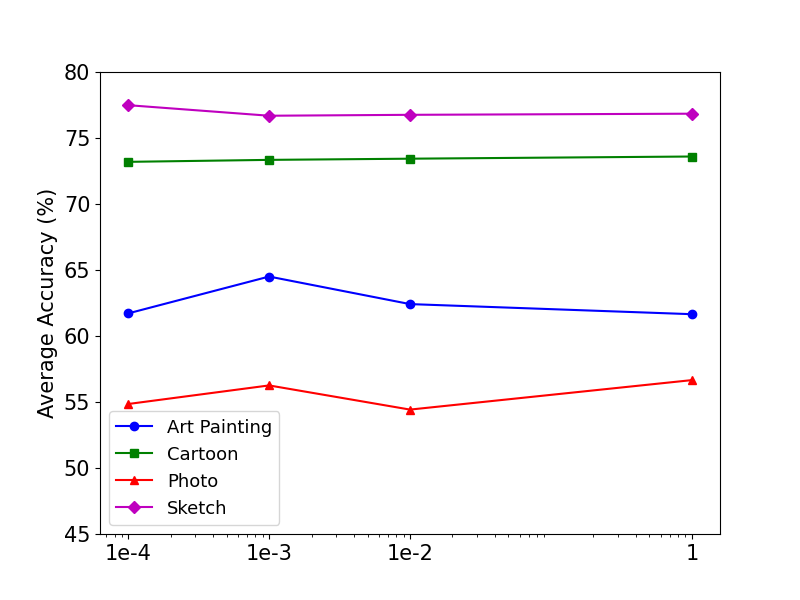}
        \caption{Parameter $\beta$ on the PACS dataset}
    \end{subfigure}
    \hfill
    \begin{subfigure}{0.23\textwidth}
        \centering
        \includegraphics[width=\textwidth]{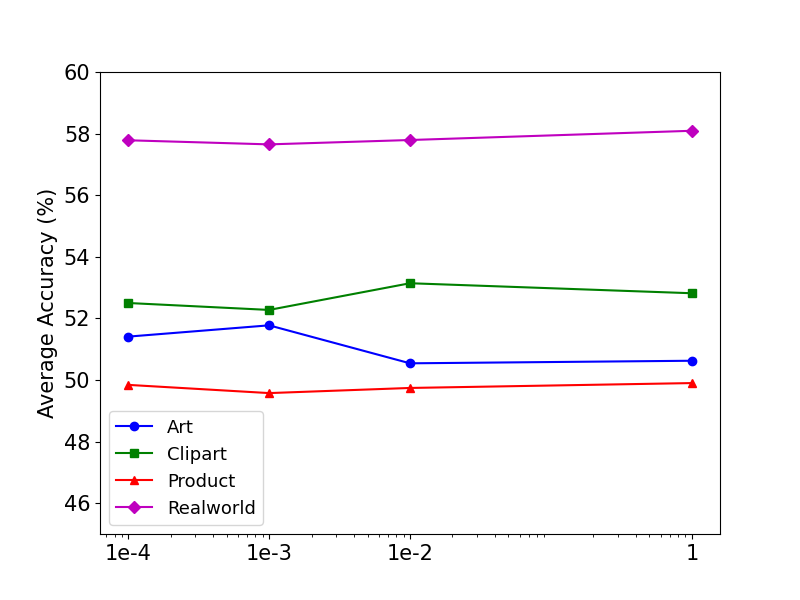}
        \caption{Parameter $\beta$ on the OfficeHome dataset}
    \end{subfigure}
    
    \vspace{0pt}

    \begin{subfigure}{0.23\textwidth}
        \centering
        \includegraphics[width=\textwidth]{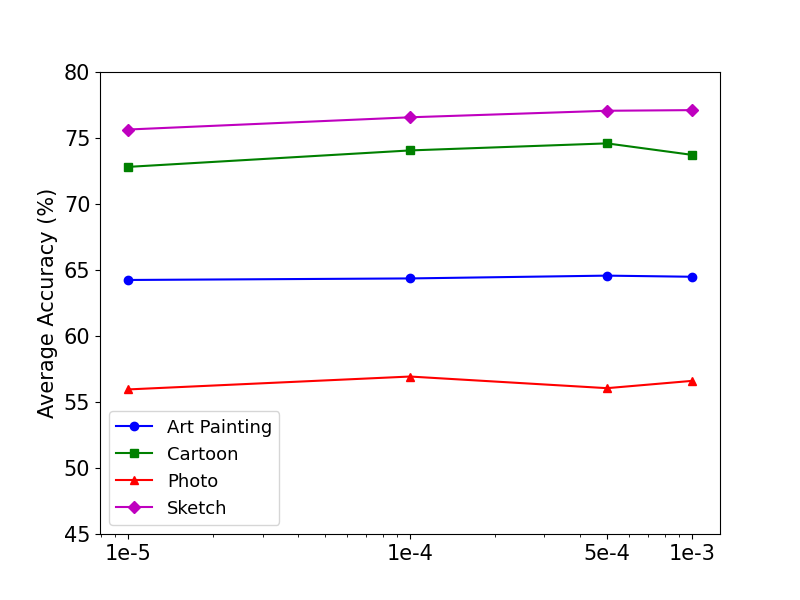}
        \caption{Parameter $\gamma$ on the PACS dataset\newline}
    \end{subfigure}
    \hfill
    \begin{subfigure}{0.23\textwidth}
        \centering
        \includegraphics[width=\textwidth]{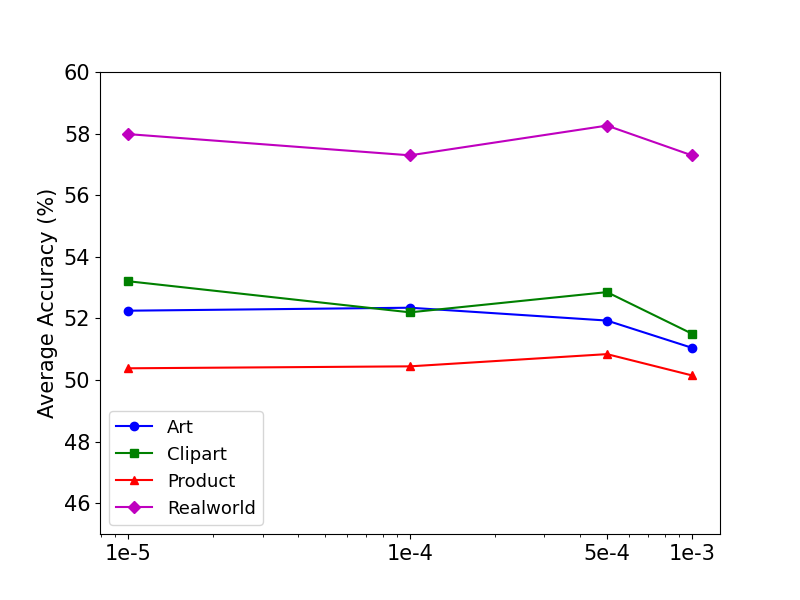}
        \caption{Parameter $\gamma$ on the OfficeHome dataset}
    \end{subfigure}
    

    \vspace{0pt} 
    \caption{Parameter sensitivity study.}
    \label{fig_parameters}
    \vspace{-10pt} 
\end{figure}

\textbf{Generalizability analysis with LEAwareSGD using different adversarial data augmentation approaches.}
To assess the impact of LEAwareSGD on the generalization performance of adversarial data augmentation methods, we integrated it with ADA, ME-ADA, and AdvST on the PACS and OfficeHome datasets. Table \ref{tab_ada_lesgd} shows consistent improvements across all methods, underscoring LEAwareSGD’s effectiveness in enhancing model generalization.
On the PACS dataset, integrating LEAwareSGD with AdvST led to a 2.40\% increase in average accuracy, reaching 69.46\%. This improvement highlights LEAwareSGD’s ability to facilitate more effective parameter space exploration, enhancing AdvST’s generalization capacity. For ADA and ME-ADA, LEAwareSGD achieved gains of 0.52\% and 2.30\%, respectively, with ME-ADA reaching an average accuracy of 62.52\%. 
On the OfficeHome dataset, LEAwareSGD also improved the performance of all methods. ME-ADA saw the highest gain, with a 7.00\% increase over ME-ADA alone, achieving an average accuracy of 52.35\%. ADA with LEAwareSGD increased by 5.15\%, reaching an average accuracy of 49.90\%. For AdvST, which already demonstrated strong performance, LEAwareSGD provided an additional 1.78\% gain, bringing the average accuracy to 54.38\%.
These findings demonstrate that LEAwareSGD enhances the generalization capacity of adversarial data augmentation techniques by enabling broader exploration of the parameter space. This exploration can facilitate learning more generalizable features across unseen domains, underscoring the effectiveness of LEAwareSGD as a robust optimization framework for SDG.


\begin{table}[htbp]
    \centering
    \setlength{\tabcolsep}{3pt}
    
    \small
    \begin{tabular}{llcccccc} 
    \toprule
        Dataset & Methods & A & C & P & S & Avg. & $\Delta$Avg. \\ \midrule
        PACS & ADA & 73.74  & 71.65  & 40.71  & 60.39  & 61.62  & ~ \\
        & ADA$^{\dagger}$ & 71.82  & 69.15  & 51.94  & 55.66  & 62.14  & +0.52  \\ 
        & ME-ADA & 71.52  & 76.83  & 46.22  & 46.32  & 60.22  & ~ \\
        & ME-ADA$^{\dagger}$ & 72.88  & 69.16  & 47.98  & 60.05  & 62.52  & +2.30 \\ 
        & AdvST & 76.65  & 74.92  & 62.47  & 54.18  & 67.06  & ~ \\
        & AdvST$^{\dagger}$  & 79.17 & 77.16 & 65.05 & 57.78 & 69.46  & +2.40 \\  
        \cmidrule(lr){2-8}
         & Methods & A & C & P & R & Avg. & $\Delta$Avg. \\ 
        \cmidrule(lr){2-8}
        \makecell{Office\\Home} & ADA & 45.07  & 36.07  & 44.43  & 53.44  & 44.75  & ~ \\
        & ADA$^{\dagger}$ & 50.69  & 48.52  & 46.11  & 54.28  & 49.90  & +5.15  \\ 
        & ME-ADA & 45.27  & 45.79  & 38.65  & 51.69  & 45.35  & ~ \\
        & ME-ADA$^{\dagger}$ & 53.61  & 50.09  & 49.17  & 56.51  & 52.35  & +7.00 \\ 
        & AdvST & 51.32  & 52.08  & 48.89  & 58.11  & 52.60  & ~ \\
        & AdvST$^{\dagger}$  &  52.89 & 55.18 & 51.69 & 58.95 & 54.38  & +1.78 \\ 
    \bottomrule
    \end{tabular}
    
    \begin{tablenotes}
    \item Note: $\dagger$ indicates methods integrated with LEAwareSGD.
\end{tablenotes}
\caption{Performance improvement (\%) of adversarial data augmentation approaches with LEAwareSGD on the PACS and OfficeHome datasets.}
    \label{tab_ada_lesgd}
\end{table}


\textbf{Parameter sensitivity analysis.} 
To assess the impact of hyperparameters on the performance of LEAwareSGD, we conducted a sensitivity analysis on the parameters beta \(\beta\) and \(\gamma\) over the PACS and OfficeHome datasets, as shown in Figure \ref{fig_parameters}. Figures \ref{fig_parameters}a and \ref{fig_parameters}b show that increasing \(\beta\) from 1e-4 to 1.0 resulted in fluctuating accuracy, with peak performance observed at $\beta=$ 1e-3 on PACS and $\beta=$ 1e-2 on OfficeHome. This result suggests an optimal range between 1e-3 and 1e-2.


As illustrated in Figure \ref{fig_parameters}c and \ref{fig_parameters}d, for the parameter \(\gamma\) in Equation \ref{eq.ada}, which controls the strength of regularization, the best accuracy on the PACS dataset was achieved at $\gamma =$ 5e-4. Increasing $\gamma$ beyond 5e-4 caused slight declines in accuracy, indicating that higher regularization may overly restrict the model. On the OfficeHome dataset, $\gamma =$ 1e-5 yielded the best accuracy across most domains, while higher values generally led to reduced performance. 

\begin{figure}[htbp]
\vspace{-10pt} 
    \centering
        \begin{subfigure}{0.23\textwidth}
        \centering
        \includegraphics[width=\textwidth]{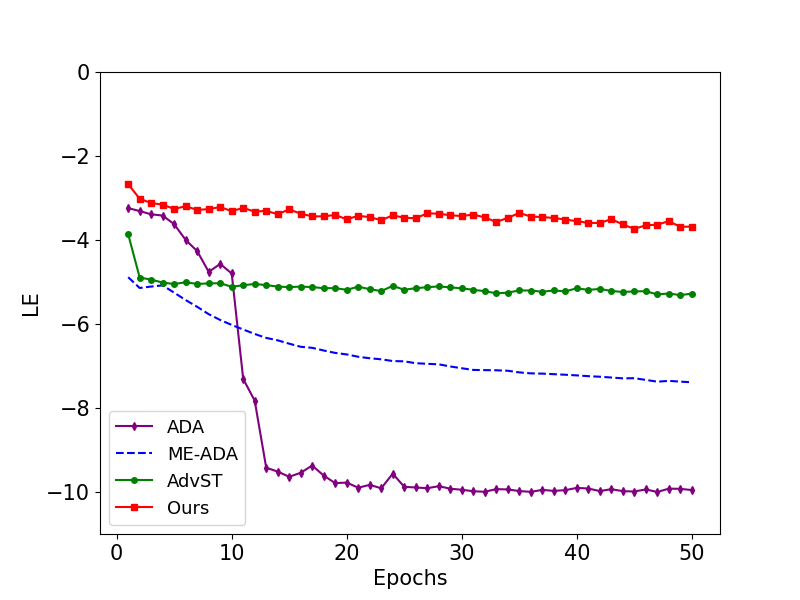}
        \caption{Trained on Art Painting domain}
    \end{subfigure}
    \hfill
    \begin{subfigure}{0.23\textwidth}
        \centering
        \includegraphics[width=\textwidth]{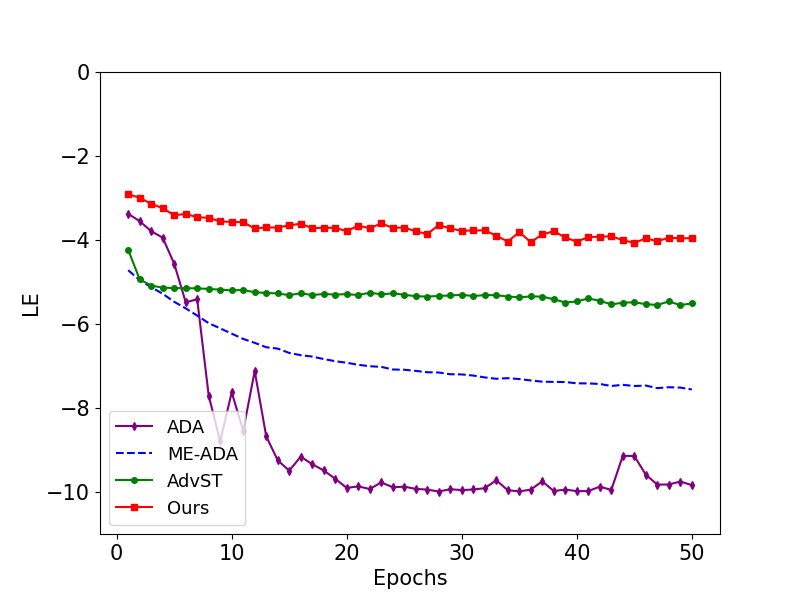}
        \caption{Trained on Cartoon domain}
    \end{subfigure}
    \vspace{0.3cm}
    \begin{subfigure}{0.23\textwidth}
        \centering
        \includegraphics[width=\textwidth]{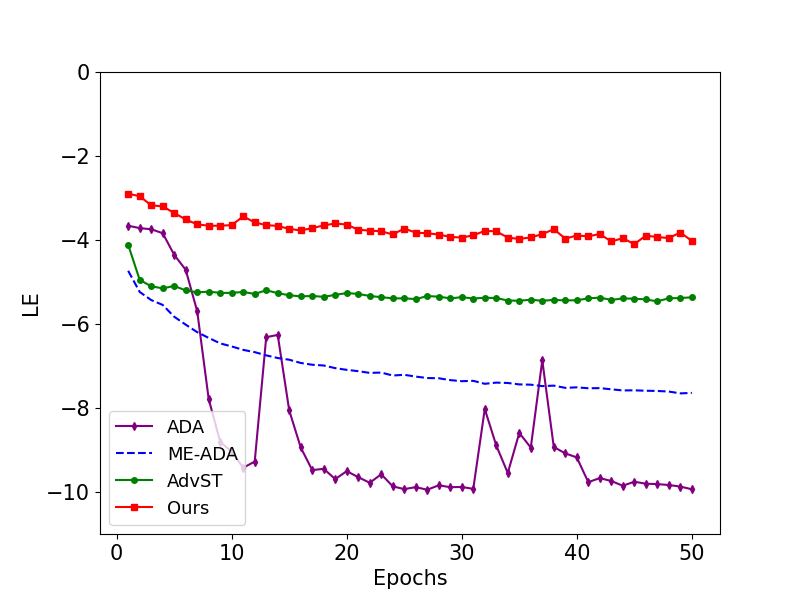}
        \caption{Trained on Photo domain}
    \end{subfigure}
    \hfill
    \begin{subfigure}{0.23\textwidth}
        \centering
        \includegraphics[width=\textwidth]{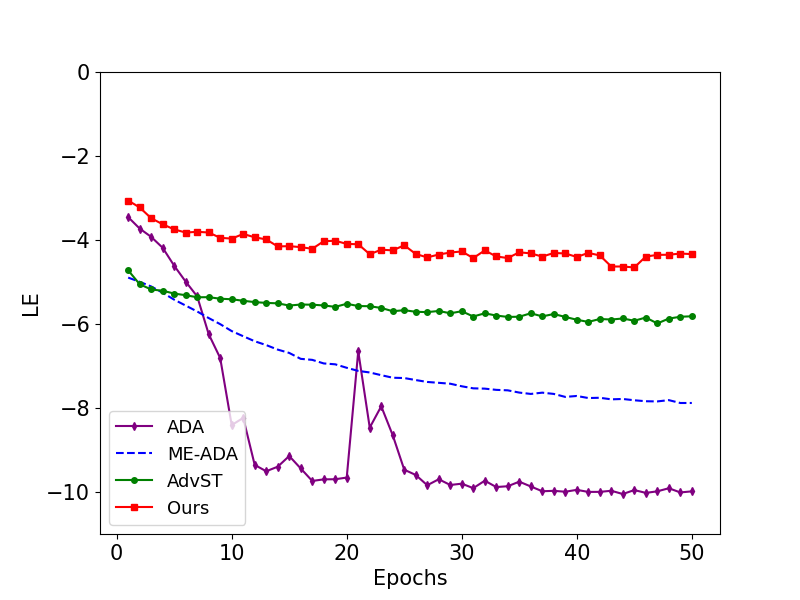}
        \caption{Trained on Sketch domain}
    \end{subfigure}
    \vspace{-10pt} 
    \caption{LE dynamics over training epochs across four domains (Art Painting, Cartoon, Photo, Sketch) on the PACS dataset.}
    \vspace{-10pt} 

    \label{fig_le}
\end{figure}


\textbf{Comparison of LE results of different methods on the PACS dataset.}
To evaluate our proposed method's stability and generalization performance, we compare the LE values obtained from different adversarial data augmentation methods on the PACS dataset. Figure \ref{fig_le} shows that we track LE dynamics across four domains:  Art Painting, Cartoon, Photo, and Sketch. Our method consistently yields LE values closer to zero, indicating that the model operates near the edge of chaos, where an optimal balance between stability and adaptability is achieved.

In the Art Painting domain (Figure \ref{fig_le}a), our method maintains LE values closer to zero throughout training compared to ADA, ME-ADA, and AdvST, suggesting a better balance between stability and adaptability, leading to improved generalization. In contrast, ADA exhibits the lowest LE values and large fluctuations, indicating instability and convergence to local optima. Similar results can be derived from the three remaining domains (Figure \ref{fig_le}b-\ref{fig_le}d), where our method consistently stabilizes near the edge of chaos, supporting better generalization. Comparison methods, by contrast, show greater fluctuations or lower LE values, indicating tendencies towards overfitting and limited adaptability. These results reveal that our method consistently achieves LE values near zero across all domains and operates effectively near the edge of chaos. 

%% file: sec/5_conclusion.tex
\section{Conclusion}
In this work, we introduce LEAwareSGD, a novel LE-guided optimization integrated with an adversarial data augmentation method to address the challenge of unseen domain shifts in SDG. By leveraging principles from dynamical systems theory, our approach encourages model training near the edge of chaos, enabling the exploration of a broader parameter space and substantially enhancing generalization to new domains. LEAwareSGD dynamically adjusts the learning rate based on LE values, encouraging the model to capture more generalizable features. Experiments on three standard SDG benchmarks demonstrate that LEAwareSGD outperforms state-of-the-art SDG methods, underscoring its effectiveness. A promising future work is to investigate its scalability to larger datasets and more complex domain generalization tasks.

%% file: main.bbl
\begin{thebibliography}{51}
\providecommand{\natexlab}[1]{#1}
\providecommand{\url}[1]{\texttt{#1}}
\expandafter\ifx\csname urlstyle\endcsname\relax
  \providecommand{\doi}[1]{doi: #1}\else
  \providecommand{\doi}{doi: \begingroup \urlstyle{rm}\Url}\fi

\bibitem[Alligood et~al.(1998)Alligood, Sauer, Yorke, and Chillingworth]{alligood1998chaos}
Kathleen~T Alligood, Tim~D Sauer, James~A Yorke, and David Chillingworth.
\newblock Chaos: an introduction to dynamical systems.
\newblock \emph{SIAM Review}, 40\penalty0 (3):\penalty0 732--732, 1998.

\bibitem[Cheng et~al.(2023)Cheng, Gokhale, and Yang]{cheng2023adversarial}
Sheng Cheng, Tejas Gokhale, and Yezhou Yang.
\newblock Adversarial bayesian augmentation for single-source domain generalization.
\newblock In \emph{Proceedings of the IEEE/CVF International Conference on Computer Vision}, pages 11400--11410, 2023.

\bibitem[Cubuk et~al.(2020)Cubuk, Zoph, Shlens, and Le]{cubuk2020randaugment}
Ekin~D Cubuk, Barret Zoph, Jonathon Shlens, and Quoc~V Le.
\newblock Randaugment: Practical automated data augmentation with a reduced search space.
\newblock In \emph{Proceedings of the IEEE/CVF Conference on Computer Vision and Pattern Recognition Workshops}, pages 702--703, 2020.

\bibitem[Cugu et~al.(2022)Cugu, Mancini, Chen, and Akata]{cugu2022attention}
Ilke Cugu, Massimiliano Mancini, Yanbei Chen, and Zeynep Akata.
\newblock Attention consistency on visual corruptions for single-source domain generalization.
\newblock In \emph{Proceedings of the IEEE/CVF Conference on Computer Vision and Pattern Recognition}, pages 4165--4174, 2022.

\bibitem[DeVries(2017)]{devries2017improved}
Terrance DeVries.
\newblock Improved regularization of convolutional neural networks with cutout.
\newblock \emph{arXiv preprint arXiv:1708.04552}, 2017.

\bibitem[Feng et~al.(2019)Feng, Zhang, and Lai]{feng2019optimal}
Ling Feng, Lin Zhang, and Choy~Heng Lai.
\newblock Optimal machine intelligence at the edge of chaos.
\newblock \emph{arXiv preprint arXiv:1909.05176}, 2019.

\bibitem[Foret et~al.(2020)Foret, Kleiner, Mobahi, and Neyshabur]{foret2020sharpness}
Pierre Foret, Ariel Kleiner, Hossein Mobahi, and Behnam Neyshabur.
\newblock Sharpness-aware minimization for efficiently improving generalization.
\newblock \emph{arXiv preprint arXiv:2010.01412}, 2020.

\bibitem[Guarneros-Sandoval et~al.(2022)Guarneros-Sandoval, Ballesteros, Salgado, Rodr{\'\i}guez-Santill{\'a}n, and Chairez]{guarneros2022lyapunov}
Alejandro Guarneros-Sandoval, Mariana Ballesteros, Ivan Salgado, Julia Rodr{\'\i}guez-Santill{\'a}n, and Isaac Chairez.
\newblock Lyapunov stable learning laws for multilayer recurrent neural networks.
\newblock \emph{Neurocomputing}, 491:\penalty0 644--657, 2022.

\bibitem[Hardt et~al.(2016)Hardt, Recht, and Singer]{hardt2016train}
Moritz Hardt, Ben Recht, and Yoram Singer.
\newblock Train faster, generalize better: Stability of stochastic gradient descent.
\newblock In \emph{International Conference on Machine Learning}, pages 1225--1234. PMLR, 2016.

\bibitem[Hayashi(2024)]{hayashi2024chaotic}
Kazuko Hayashi.
\newblock Chaotic nature of the electroencephalogram during shallow and deep anesthesia: From analysis of the lyapunov exponent.
\newblock \emph{Neuroscience}, 557:\penalty0 116--123, 2024.

\bibitem[He et~al.(2016)He, Zhang, Ren, and Sun]{he2016deep}
Kaiming He, Xiangyu Zhang, Shaoqing Ren, and Jian Sun.
\newblock Deep residual learning for image recognition.
\newblock In \emph{Proceedings of the IEEE Conference on Computer Vision and Pattern Recognition}, pages 770--778, 2016.

\bibitem[Hendrycks et~al.(2019)Hendrycks, Mu, Cubuk, Zoph, Gilmer, and Lakshminarayanan]{hendrycks2019augmix}
Dan Hendrycks, Norman Mu, Ekin~D Cubuk, Barret Zoph, Justin Gilmer, and Balaji Lakshminarayanan.
\newblock Augmix: A simple data processing method to improve robustness and uncertainty.
\newblock \emph{arXiv preprint arXiv:1912.02781}, 2019.

\bibitem[Huang et~al.(2020)Huang, Wang, Xing, and Huang]{huang2020self}
Zeyi Huang, Haohan Wang, Eric~P Xing, and Dong Huang.
\newblock Self-challenging improves cross-domain generalization.
\newblock In \emph{ECCV}, pages 124--140, 2020.

\bibitem[Kingma(2014)]{kingma2014adam}
Diederik~P Kingma.
\newblock Adam: A method for stochastic optimization.
\newblock \emph{arXiv preprint arXiv:1412.6980}, 2014.

\bibitem[Li et~al.(2017)Li, Yang, Song, and Hospedales]{li2017deeper}
Da Li, Yongxin Yang, Yi-Zhe Song, and Timothy~M Hospedales.
\newblock Deeper, broader and artier domain generalization.
\newblock In \emph{Proceedings of the IEEE International Conference on Computer Vision}, pages 5542--5550, 2017.

\bibitem[Li et~al.(2021)Li, Gao, Cao, Huang, Weng, Mi, Yu, Li, and Xia]{li2021progressive}
Lei Li, Ke Gao, Juan Cao, Ziyao Huang, Yepeng Weng, Xiaoyue Mi, Zhengze Yu, Xiaoya Li, and Boyang Xia.
\newblock Progressive domain expansion network for single domain generalization.
\newblock In \emph{Proceedings of the IEEE/CVF Conference on Computer Vision and Pattern Recognition}, pages 224--233, 2021.

\bibitem[Li et~al.(2022)Li, Lin, and Shen]{li2022deep}
Qianxiao Li, Ting Lin, and Zuowei Shen.
\newblock Deep learning via dynamical systems: An approximation perspective.
\newblock \emph{Journal of the European Mathematical Society}, 25\penalty0 (5):\penalty0 1671--1709, 2022.

\bibitem[Li et~al.(2019)Li, Song, Fang, Chen, Ghamisi, and Benediktsson]{li2019deep}
Shutao Li, Weiwei Song, Leyuan Fang, Yushi Chen, Pedram Ghamisi, and Jon~Atli Benediktsson.
\newblock Deep learning for hyperspectral image classification: An overview.
\newblock \emph{IEEE Transactions on Geoscience and Remote Sensing}, 57\penalty0 (9):\penalty0 6690--6709, 2019.

\bibitem[Liu and Theodorou(2019)]{liu2019deep}
Guan-Horng Liu and Evangelos~A Theodorou.
\newblock Deep learning theory review: An optimal control and dynamical systems perspective.
\newblock \emph{arXiv preprint arXiv:1908.10920}, 2019.

\bibitem[Loshchilov(2017)]{loshchilov2017decoupled}
I Loshchilov.
\newblock Decoupled weight decay regularization.
\newblock \emph{arXiv preprint arXiv:1711.05101}, 2017.

\bibitem[Nam et~al.(2021)Nam, Lee, Park, Yoon, and Yoo]{nam2021reducing}
Hyeonseob Nam, HyunJae Lee, Jongchan Park, Wonjun Yoon, and Donggeun Yoo.
\newblock Reducing domain gap by reducing style bias.
\newblock In \emph{Proceedings of the IEEE/CVF Conference on Computer Vision and Pattern Recognition}, pages 8690--8699, 2021.

\bibitem[Nesterov(1983)]{nesterov1983method}
Yurii Nesterov.
\newblock A method for solving the convex programming problem with convergence rate o (1/k2).
\newblock In \emph{Proceedings of the USSR Academy of Sciences}, pages 543--547, 1983.

\bibitem[Pan and Yang(2009)]{pan2009survey}
Sinno~Jialin Pan and Qiang Yang.
\newblock A survey on transfer learning.
\newblock \emph{IEEE Transactions on Knowledge and Data Engineering}, 22\penalty0 (10):\penalty0 1345--1359, 2009.

\bibitem[Peng et~al.(2019)Peng, Bai, Xia, Huang, Saenko, and Wang]{peng2019moment}
Xingchao Peng, Qinxun Bai, Xide Xia, Zijun Huang, Kate Saenko, and Bo Wang.
\newblock Moment matching for multi-source domain adaptation.
\newblock In \emph{Proceedings of the IEEE/CVF International Conference on Computer Vision}, pages 1406--1415, 2019.

\bibitem[Qiao et~al.(2020)Qiao, Zhao, and Peng]{qiao2020learning}
Fengchun Qiao, Long Zhao, and Xi Peng.
\newblock Learning to learn single domain generalization.
\newblock In \emph{Proceedings of the IEEE/CVF Conference on Computer Vision and Pattern Recognition}, pages 12556--12565, 2020.

\bibitem[Robbins and Monro(1951)]{robbins1951stochastic}
Herbert Robbins and Sutton Monro.
\newblock A stochastic approximation method.
\newblock \emph{The annals of mathematical statistics}, pages 400--407, 1951.

\bibitem[Shao et~al.(2014)Shao, Zhu, and Li]{shao2014transfer}
Ling Shao, Fan Zhu, and Xuelong Li.
\newblock Transfer learning for visual categorization: A survey.
\newblock \emph{IEEE Transactions on Neural Networks and Learning Systems}, 26\penalty0 (5):\penalty0 1019--1034, 2014.

\bibitem[Shi et~al.(2021)Shi, Seely, Torr, Siddharth, Hannun, Usunier, and Synnaeve]{shi2021gradient}
Yuge Shi, Jeffrey Seely, Philip~HS Torr, N Siddharth, Awni Hannun, Nicolas Usunier, and Gabriel Synnaeve.
\newblock Gradient matching for domain generalization.
\newblock \emph{arXiv preprint arXiv:2104.09937}, 2021.

\bibitem[Tang et~al.(2020)Tang, Kurths, Lin, Ott, and Kocarev]{tang2020introduction}
Yang Tang, J{\"u}rgen Kurths, Wei Lin, Edward Ott, and Ljupco Kocarev.
\newblock Introduction to focus issue: When machine learning meets complex systems: Networks, chaos, and nonlinear dynamics.
\newblock \emph{Chaos: An Interdisciplinary Journal of Nonlinear Science}, 30\penalty0 (6), 2020.

\bibitem[Tieleman and Hinton(2012)]{tieleman2012rmsprop}
Tijmen Tieleman and Geoffrey Hinton.
\newblock Rmsprop: Divide the gradient by a running average of its recent magnitude. coursera: Neural networks for machine learning.
\newblock \emph{COURSERA Neural Networks Mach. Learn}, 17, 2012.

\bibitem[Venkateswara et~al.(2017)Venkateswara, Eusebio, Chakraborty, and Panchanathan]{venkateswara2017deep}
Hemanth Venkateswara, Jose Eusebio, Shayok Chakraborty, and Sethuraman Panchanathan.
\newblock Deep hashing network for unsupervised domain adaptation.
\newblock In \emph{Proceedings of the IEEE Conference on Computer Vision and Pattern Recognition}, pages 5018--5027, 2017.

\bibitem[Vogt et~al.(2022)Vogt, Puelma~Touzel, Shlizerman, and Lajoie]{vogt2022lyapunov}
Ryan Vogt, Maximilian Puelma~Touzel, Eli Shlizerman, and Guillaume Lajoie.
\newblock On lyapunov exponents for rnns: Understanding information propagation using dynamical systems tools.
\newblock \emph{Frontiers in Applied Mathematics and Statistics}, 8:\penalty0 818799, 2022.

\bibitem[Volpi et~al.(2018)Volpi, Namkoong, Sener, Duchi, Murino, and Savarese]{volpi2018generalizing}
Riccardo Volpi, Hongseok Namkoong, Ozan Sener, John~C Duchi, Vittorio Murino, and Silvio Savarese.
\newblock Generalizing to unseen domains via adversarial data augmentation.
\newblock In \emph{Proceedings of the 32nd International Conference on Neural Information Processing Systems}, pages 5339--5349, 2018.

\bibitem[Wan et~al.(2022)Wan, Shen, Zhang, Yin, Tian, Gao, Huang, and Hua]{wan2022meta}
Chaoqun Wan, Xu Shen, Yonggang Zhang, Zhiheng Yin, Xinmei Tian, Feng Gao, Jianqiang Huang, and Xian-Sheng Hua.
\newblock Meta convolutional neural networks for single domain generalization.
\newblock In \emph{Proceedings of the IEEE/CVF Conference on Computer Vision and Pattern Recognition}, pages 4682--4691, 2022.

\bibitem[Wang et~al.(2023)Wang, Zhang, Lei, and Zhang]{wang2023sharpness}
Pengfei Wang, Zhaoxiang Zhang, Zhen Lei, and Lei Zhang.
\newblock Sharpness-aware gradient matching for domain generalization.
\newblock In \emph{Proceedings of the IEEE/CVF Conference on Computer Vision and Pattern Recognition}, pages 3769--3778, 2023.

\bibitem[Wang et~al.(2021)Wang, Luo, Qiu, Huang, and Baktashmotlagh]{wang2021learning}
Zijian Wang, Yadan Luo, Ruihong Qiu, Zi Huang, and Mahsa Baktashmotlagh.
\newblock Learning to diversify for single domain generalization.
\newblock In \emph{Proceedings of the IEEE/CVF International Conference on Computer Vision}, pages 834--843, 2021.

\bibitem[Xu et~al.(2023)Xu, Zhang, Wu, Zhang, Liu, and Wang]{xu2023simde}
Qinwei Xu, Ruipeng Zhang, Yi-Yan Wu, Ya Zhang, Ning Liu, and Yanfeng Wang.
\newblock Simde: A simple domain expansion approach for single-source domain generalization.
\newblock In \emph{Proceedings of the IEEE/CVF Conference on Computer Vision and Pattern Recognition}, pages 4798--4808, 2023.

\bibitem[Xu et~al.(2020)Xu, Liu, Yang, Raffel, and Niethammer]{xu2020robust}
Zhenlin Xu, Deyi Liu, Junlin Yang, Colin Raffel, and Marc Niethammer.
\newblock Robust and generalizable visual representation learning via random convolutions.
\newblock \emph{arXiv preprint arXiv:2007.13003}, 2020.

\bibitem[Yan et~al.(2020)Yan, Song, Li, Zou, and Ren]{yan2020improve}
Shen Yan, Huan Song, Nanxiang Li, Lincan Zou, and Liu Ren.
\newblock Improve unsupervised domain adaptation with mixup training.
\newblock \emph{arXiv preprint arXiv:2001.00677}, 2020.

\bibitem[Yang et~al.(2024{\natexlab{a}})Yang, Zhang, and Gu]{yang2024causality}
Shuai Yang, Zhen Zhang, and Lichuan Gu.
\newblock Causality-inspired domain expansion network for single domain generalization.
\newblock \emph{Knowledge-Based Systems}, 301:\penalty0 112269, 2024{\natexlab{a}}.

\bibitem[Yang et~al.(2024{\natexlab{b}})Yang, Zhang, and Gu]{yang2024practical}
Shuai Yang, Zhen Zhang, and Lichuan Gu.
\newblock Practical single domain generalization via training-time and test-time learning.
\newblock In \emph{Proceedings of the 30th ACM SIGKDD Conference on Knowledge Discovery and Data Mining}, pages 3794--3805, 2024{\natexlab{b}}.

\bibitem[Yu et~al.(2020)Yu, Kumar, Gupta, Levine, Hausman, and Finn]{yu2020gradient}
Tianhe Yu, Saurabh Kumar, Abhishek Gupta, Sergey Levine, Karol Hausman, and Chelsea Finn.
\newblock Gradient surgery for multi-task learning.
\newblock \emph{Advances in Neural Information Processing Systems}, 33:\penalty0 5824--5836, 2020.

\bibitem[Zhang et~al.(2020)Zhang, Wang, Yang, Sanford, Harmon, Turkbey, Wood, Roth, Myronenko, Xu, et~al.]{zhang2020generalizing}
Ling Zhang, Xiaosong Wang, Dong Yang, Thomas Sanford, Stephanie Harmon, Baris Turkbey, Bradford~J Wood, Holger Roth, Andriy Myronenko, Daguang Xu, et~al.
\newblock Generalizing deep learning for medical image segmentation to unseen domains via deep stacked transformation.
\newblock \emph{IEEE Transactions on Medical Imaging}, 39\penalty0 (7):\penalty0 2531--2540, 2020.

\bibitem[Zhang et~al.(2021)Zhang, Feng, Chen, and Lai]{zhang2021edge}
Lin Zhang, Ling Feng, Kan Chen, and Choy~Heng Lai.
\newblock Edge of chaos as a guiding principle for modern neural network training.
\newblock \emph{arXiv preprint arXiv:2107.09437}, 2021.

\bibitem[Zhang et~al.(2024)Zhang, Feng, Chen, and Lai]{zhang2024asymptotic}
Lin Zhang, Ling Feng, Kan Chen, and Choy~Heng Lai.
\newblock Asymptotic edge of chaos as guiding principle for neural network training.
\newblock \emph{International Journal of Artificial Intelligence and Robotics Research}, 1\penalty0 (01):\penalty0 2350001, 2024.

\bibitem[Zhang et~al.(2023)Zhang, Xu, Yu, Dong, Tian, and Cui]{zhang2023flatness}
Xingxuan Zhang, Renzhe Xu, Han Yu, Yancheng Dong, Pengfei Tian, and Peng Cui.
\newblock Flatness-aware minimization for domain generalization.
\newblock In \emph{Proceedings of the IEEE/CVF International Conference on Computer Vision}, pages 5189--5202, 2023.

\bibitem[Zhao et~al.(2020)Zhao, Liu, Peng, and Metaxas]{zhao2020maximum}
Long Zhao, Ting Liu, Xi Peng, and Dimitris Metaxas.
\newblock Maximum-entropy adversarial data augmentation for improved generalization and robustness.
\newblock \emph{Advances in Neural Information Processing Systems}, 33:\penalty0 14435--14447, 2020.

\bibitem[Zheng et~al.(2024)Zheng, Huai, and Zhang]{zheng2024advst}
Guangtao Zheng, Mengdi Huai, and Aidong Zhang.
\newblock Advst: Revisiting data augmentations for single domain generalization.
\newblock In \emph{Proceedings of the AAAI Conference on Artificial Intelligence}, pages 21832--21840, 2024.

\bibitem[Zhou et~al.(2024)Zhou, Yang, Qiao, and Xiang]{zhou2024mixstyle}
Kaiyang Zhou, Yongxin Yang, Yu Qiao, and Tao Xiang.
\newblock Mixstyle neural networks for domain generalization and adaptation.
\newblock \emph{International Journal of Computer Vision}, 132\penalty0 (3):\penalty0 822--836, 2024.

\bibitem[Zhuang et~al.(2020)Zhuang, Qi, Duan, Xi, Zhu, Zhu, Xiong, and He]{zhuang2020comprehensive}
Fuzhen Zhuang, Zhiyuan Qi, Keyu Duan, Dongbo Xi, Yongchun Zhu, Hengshu Zhu, Hui Xiong, and Qing He.
\newblock A comprehensive survey on transfer learning.
\newblock \emph{Proceedings of the IEEE}, 109\penalty0 (1):\penalty0 43--76, 2020.

\bibitem[Zhuang et~al.(2022)Zhuang, Gong, Yuan, Cui, Adam, Dvornek, Tatikonda, Duncan, and Liu]{zhuang2022surrogate}
Juntang Zhuang, Boqing Gong, Liangzhe Yuan, Yin Cui, Hartwig Adam, Nicha Dvornek, Sekhar Tatikonda, James Duncan, and Ting Liu.
\newblock Surrogate gap minimization improves sharpness-aware training.
\newblock \emph{arXiv preprint arXiv:2203.08065}, 2022.

\end{thebibliography}
